\documentclass{article}
\usepackage{adjustbox,multirow}
\usepackage{spconf,amsmath,graphicx}
\usepackage{hyperref}
\usepackage{mathtools}
\DeclarePairedDelimiter{\ceil}{\lceil}{\rceil}

\title{Multi-Patch Aggregation Models for Resampling Detection}
\name{Mohit Lamba \qquad Kaushik Mitra}
\address{Department of Electrical Engineering, Indian Institute of Technology Madras}
\def\etal{\emph{et al.}}
\begin{document}
%\ninept
%
\maketitle
\begin{abstract}
Images captured nowadays are of varying dimensions with smartphones and DSLR's allowing users to choose from a list of available image resolutions. It is therefore imperative for forensic algorithms such as resampling detection to scale well for images of varying dimensions. However, in our experiments we observed that many state-of-the-art forensic algorithms are sensitive to image size and their performance quickly degenerates when operated on images of diverse dimensions despite re-training them using multiple image sizes. To handle this issue, we propose two novel deep neural networks \textemdash  ~\textit{Iterative Pooling Network (IPN)}, which does not assume any prior information about the original image size, and \textit{Branched Network (BN)}, which uses this prior knowledge to produce better results. \textit{IPN} adopts a novel \textit{iterative pooling} strategy that converts tensors of multiple sizes to tensors of a fixed size, as required by deep learning models with fully connected layers. \textit{BN} alternatively adopts a branched architecture with dedicated pathways for images of different sizes.
The effectiveness of the proposed solution is demonstrated on two problems, resampling detection and photorealism detection, which are generally solved as independent problems with different deep learning models. The code is available at \url{https://github.com/MohitLamba94/Iterative-Pooling}.
%Leveraging on existing pre-trained deep learning models has helped several vision tasks such as object detection to obtain state-of-the-art performance in terms of speed and accuracy. This saves from meticulously choosing new design parameters with efficient implementations readily available for use. In this work, we show that using pre-trained ResNet model achieves at-par accuracy with faster convergence for much later proposed forensics models. Also, the performance of existing models is limited to images of similar size. Their performance quickly degenerates for images of varying dimensions which is quite common nowadays. We, therefore, modify the pre-trained ResNet architecture to work with patches of different sizes to handle images of varying dimensions. This saves processing time and the need for re-training models for different dimensions. The effectiveness of the proposed model is demonstrated on two problems, namely resampling detection with double JPEG compression and photorealism detection of heterogeneous origin, which generally are solved as independent problems with different model choices.
\end{abstract}
\section{Introduction}
\label{sec:intro}

Resampling detection has been one of the extensively studied topics because any type of image manipulation often involves this operation \cite{farid2005resampling, bianchiPiva2012resamplingJPEG,welchPSD,2017stamm}. The earlier works focused on uncompressed images \cite{farid2005resampling} or considered just upsampling \cite{gallagher} but a more realistic scenario would involve double compressed images \cite{bianchiPiva2012resamplingJPEG, welchPSD}, once at the time of acquisition and then again while saving the image after manipulation. The case of compressed images with resampling is complicated because it results in non-aligned compression \cite{piva2013overview} where compression traits camouflage with resampling traits. More critical however is the fact that many compression techniques such as JPEG2000 can tile the image in blocks of arbitrary sizes. In many standard implementations, such as MATLAB JPEG2000 encoder, this block size is as big as the entire image. Thus, for detecting resampling the input patch size should be chosen based on the image size. However, this is not possible for neural networks with fully connected (\textit{fc}) layers \cite{2017stamm, stamm2018TIFS}, the most popular choice for classification nowadays.

\begin{figure}[t!]
    \centering
    \includegraphics[width=\linewidth]{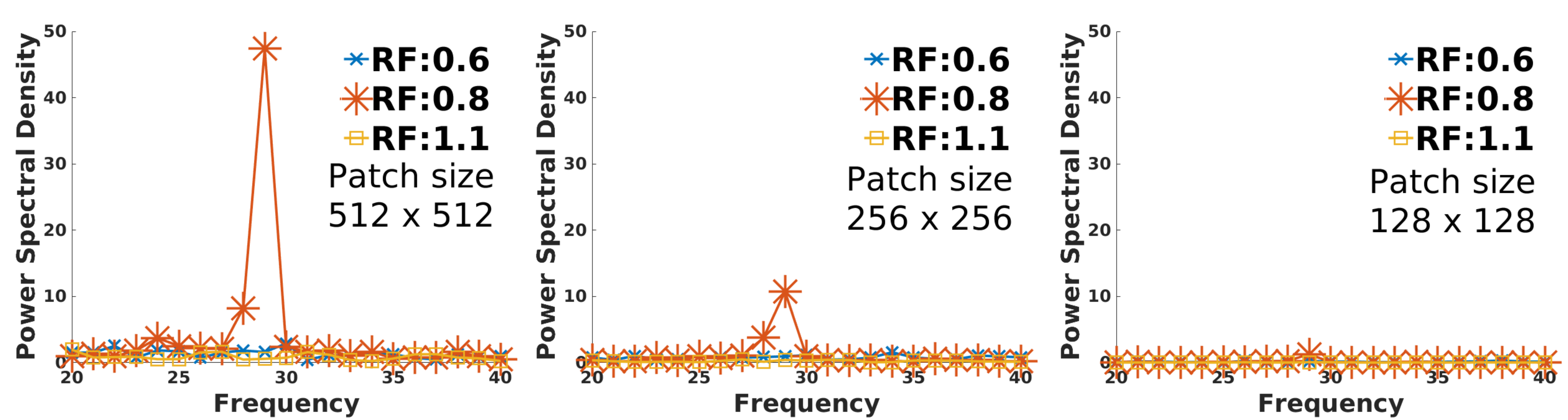}
    \caption{Resampling factor (RF) estimation, using a recently proposed method by Sahu and Okade \cite{welchPSD}, for a double JPEG compressed image of size $1024\times1024$ and resampled by $0.8$. The performance is evaluated on 3 patch sizes. Since the accuracy is contingent on the sharpness of peak, it is obsevred the performance of the method quickly degrades for patches less than  $512 \times 512$. }
    \label{welsh_ineffective}
\end{figure}

Another problem with operating on patches of fixed size, say $128\times128$, is that it may be the optimum choice for images of dimension, say $512\times512$, but it is too small for images of larger dimensions, say $4000\times4000$. This is illustrated in Fig. \ref{welsh_ineffective}. On the other hand one can always choose a large patch size for better performance. But the downside of this choice is that it won't work for smaller image sizes and the algorithm would incur excessive computations and processing time and poor manipulation localization \cite{hemant, multJPEG}. It is perhaps for this reason that many recent works have considered images of roughly the same size and consequently a fixed patch size. Refer to Table \ref{earlier_base} for some of the recent works with their choice of image and patch sizes. Our experiments show that the performance of these algorithms quickly degenerate when operated on images of varying dimensions.
%Perhaps some of the works based on statistical features such as \cite{welchPSD,bianchiPiva2012resamplingJPEG} work well with only a large patch size of $512\times512$ or more. 

%Working with multiple patch sizes, however, has traditionally not been possible for neural networks with fully connected layers, the most popular choice for detection nowadays. 
 
%This is perhaps the main reason for recent neural network based solutions \cite{stamm2018TIFS, 2017stamm, tifs2018prcg, multJPEG} to work with uniform patch sizes.  In a realistic scenario, however, these algorithms will have to operate on images of varying sizes requiring them to alter the patch sizes accordingly. 

To handle the issue of variable image size, we propose two deep neural networks \textemdash~ \textit{Iterative Pooling Network (IPN)}, that assumes no prior information about the original image size and \textit{Branched Network (BN)} that assumes some information about the same. \textit{IPN} integrates the proposed \textit{iterative pooling} strategy to existing deep neural networks which transforms variable patch sizes to fixed-size tensors before passing them to the \textit{fc} layer of the network.  \textit{BN}, on the other hand, requires coarse level information about image sizes to separately process images of different size for better performance. We show that by leveraging on pre-trained models, the fine-tuning of the proposed networks not only converges much faster than recent works such as \cite{2017stamm, stamm2018TIFS} but also generalizes well for other tasks like photorealism detection of heterogeneous origin \cite{tifs2018prcg}. 
%These two problems have been solved independently to date in the existing literature.
%One possible solution would be to maintain an ensemble of many such classifiers, each trained for images of similar sizes, but this may be prohibitive due to time and computation constraints and is rather an inefficient way to handle the problem. 
%It is beneficial to operate on small patch sizes which exceedingly reduces the computation and time complexity and is at times useful for localization by tiling the classifier over the entire image. Using small patch sizes, many times also helps to train the classifier with significantly less training data.

\section{Iterative Pooling}
Patches of different sizes result in feature maps (tensors) of different sizes at various layers of a neural network. This is alright for convolution layers which can operate on tensors of any height and width but it is problematic for the fully connected layers. One possible solution, as adopted by Faster RCNN \cite{fasterRCNN}, is to max pool \cite{maxpooling,Alexnet} the incoming tensor with kernel size proportional to tensor's height and width to reduce all tensors to the same size. However, in our experiments this resulted in poor performance, see results for the Max-Pooling Network (MPN) in Table \ref{table:mainn}. This is most likely due to the heavy loss of information which happens due to max-pooling. As an illustratation, assume that the the incoming tensor is of size $H\times H \times C$ and the expected output dimension is $h \times h \times C$. Then the number of data points discarded in max-pooling is given by $(H^2 - h^2)C$ with one data point retained in every $\frac{H^2}{h^2} \times \frac{H^2}{h^2} \times 1$ sub-block.
Hence, if $H=16$, $C=128$ and $h=4$, the number of data points discarded is around $40k$. 
%In contrast, if $H=64$, more than $500~thousand$ data points are discarded. 
To avoid this heavy information loss, we propose a new pooling strategy called \textit{iterative pooling} which significantly boosts the classification accuracy by discretizing the space of patch dimensions but with minimal loss in information. \textit{Iterative pooling} downsizes the input tensor by iteratively convolving it with a $3 \times 3 \times C$ kernel with a stride of $2$ for $\log_2 \left(\ceil{\frac{H}{h}}\right)$ iterations %($\ceil{x}$ denotes rounding to the nearest integer greater than $x$). 
The convolution weights are shared across the patch sizes and across the time steps. In each time step the height and width of the tensor reduces by half. Since the operation is carried out for logarithm time steps, with $\frac{H}{h}$ being a small number, the operation completes almost instantaneously.

\begin{table}[t!]
\centering
\caption{Recent works with image size in their dataset and the corresponding patch size.}
\begin{tabular}{ccc}
\hline \textbf{Work}   & \textbf{Image Size} & \textbf{Patch Size} \\ \hline
Sahu \etal~\cite{welchPSD}    & $1024\times1024$                & $512\times512$             \\
MISLnet~\cite{stamm2018TIFS}         & $256\times256$                  & $256\times256$             \\
Li \etal~ \cite{variousImageManipulation} & $512\times512$ & $512\times512$\\
Verma \etal~ \cite{multJPEG}    & $512\times384$                  & $128\times128$             \\
Kirchner \etal~ \cite{kirchner2009resampling} & $1024\times1024$                & $512\times512$             \\
Bianchi \etal~ \cite{bianchiPiva2012resamplingJPEG}   & $1024\times1024$                & $512\times512$             \\
%Amerini \etal~ \cite{patch64} & $512\times384$                  & $64\times64$               \\
Quan \etal~ \cite{tifs2018prcg}    & \textless{}$1024\times1024$  & $233\times233$ \\ \hline          
\end{tabular}

\label{earlier_base}
\end{table}

%Let, the image patch $I \in \mathbf{R}^{H \times H \times 3}$ with $H$ taking only discrete values, namely  $H\in\{64,128, 256, 512, 1024\}$. This discretization of the continuous variable $H$ is however, not a serious limitation for the classification task. The given patch sizes suffice for images of various dimensions and are suitable approximations of the desired value of $H$. For example, it is very unlikely that the performance with a patch size of $300\times300$ should outperform the performance of a deep neural network with a patch size of $256\times256$. 

%$I$ first passes through various convolution layers of a neural network, denoted by $F_1(I) \in \mathbf{R}^{H/m \times H/m \times C}$, which is then downsized to a fixed size using the iterative pooling strategy, denoted by $pooling$. Let, $pooling(F_1(I))\in\mathbf{R}^{h\times h\times C}$, where $h$ is independent of the dimension of $F_1(I)$ and is fixed at $4$ in our experiments. $pooling(F_1(I))$ is then processed by the fully connected layers for the final classification.

Inspired by the immense success of works like Faster RCNN \cite{fasterRCNN} and YOLO \cite{yolo} which decided to build on top of existing CNN models, we also chose to use ResNet-18 \cite{ResidualBlock} as our base architecture. Although proposed much earlier than recent CNN models for image forensic \cite{2017stamm, stamm2018TIFS, tifs2018prcg} we found that ResNet-18 with our iterative pooling strategy produces much better results. Also, the pre-trained ResNet-18 model significantly improves the convergence time of our algorithm. We inserted the proposed \textit{iterative pooling} strategy in between the $conv3\_x \text{ and } conv4\_x$ layers of ResNet-18 to handle patches of varying dimension. In our experiments, $h$ was fixed at $4$. 

\begin{figure}[t!]
    \centering
    \includegraphics[width = 0.9\linewidth]{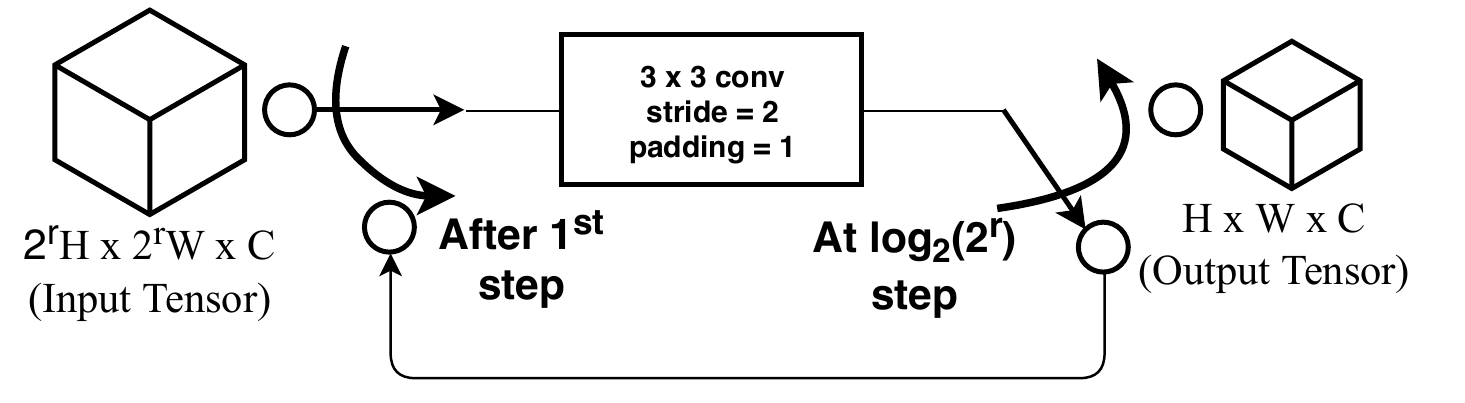}
    \caption{A pictorial representation of the proposed \textit{iterative pooling} strategy that is integrated into the proposed deep neural network, Iterative Pooling Network (IPN), to handle patches of multiple sizes which was not possible traditionally. Despite its simplicity it helps achieves much higher accuracy as confirmed in our experiments.}
    \label{fig:iter_pooling}
\end{figure}

\textit{Iterative pooling} can result in tensors with a non-integer dimension requiring appropriate padding before convolution. This problem is also common to naive max-pooling strategy discussed above. To avoid this, we discretize the input patch space and accept patch sizes which can be expressed as some integer power of two, i.e., $2^r$, see Fig. \ref{fig:iter_pooling}. This discretization is not a serious limitation for the classification task. The given patch sizes suffice for images of various dimensions and are suitable approximations of the desired value of patch dimension. For example, it is very unlikely that the performance with a patch size of $300\times300$ should outperform the performance of a deep neural network with a patch size of $256\times256$.
We call this architecture \textit{Iterative Pooling Network (IPN)}.
%To check for the effectiveness of \textit{iterative pooling}, we also show results for Max Pooling Network (MPN) which uses max pooling with adaptive kernel size instead of \textit{iterative pooling} in IPN (Refer Table \ref{table:mainn}). 

Motivated by the idea of using a recursive block in between an otherwise feedforward neural network, we propose another solution in the next section called \textit{Branched Network (BN)} which further helps to boost the accuracy provided some prior information about the original image size is given to the network.
\label{sec: iterative pooling}

\section{Branched Network}
In some forensic applications such as dealing with CCTV footage we have a coarse idea of the image size. This idea was however not harnessed in the previous section and we shall show in our experiments that utilizing this information further helps to boost the performance. We divide the images into three categories, namely, category I, II and III for small, medium and large-sized images respectively. These image sizes refer to image dimensions before resampling. Fixed patch size is used for each category irrespective of the resampling factor. Category I uses a patch size of $64\times64$, Category II of $128\times128$ and Category III of $256\times256$. 

\begin{figure}
    \centering
    \includegraphics[scale=0.5]{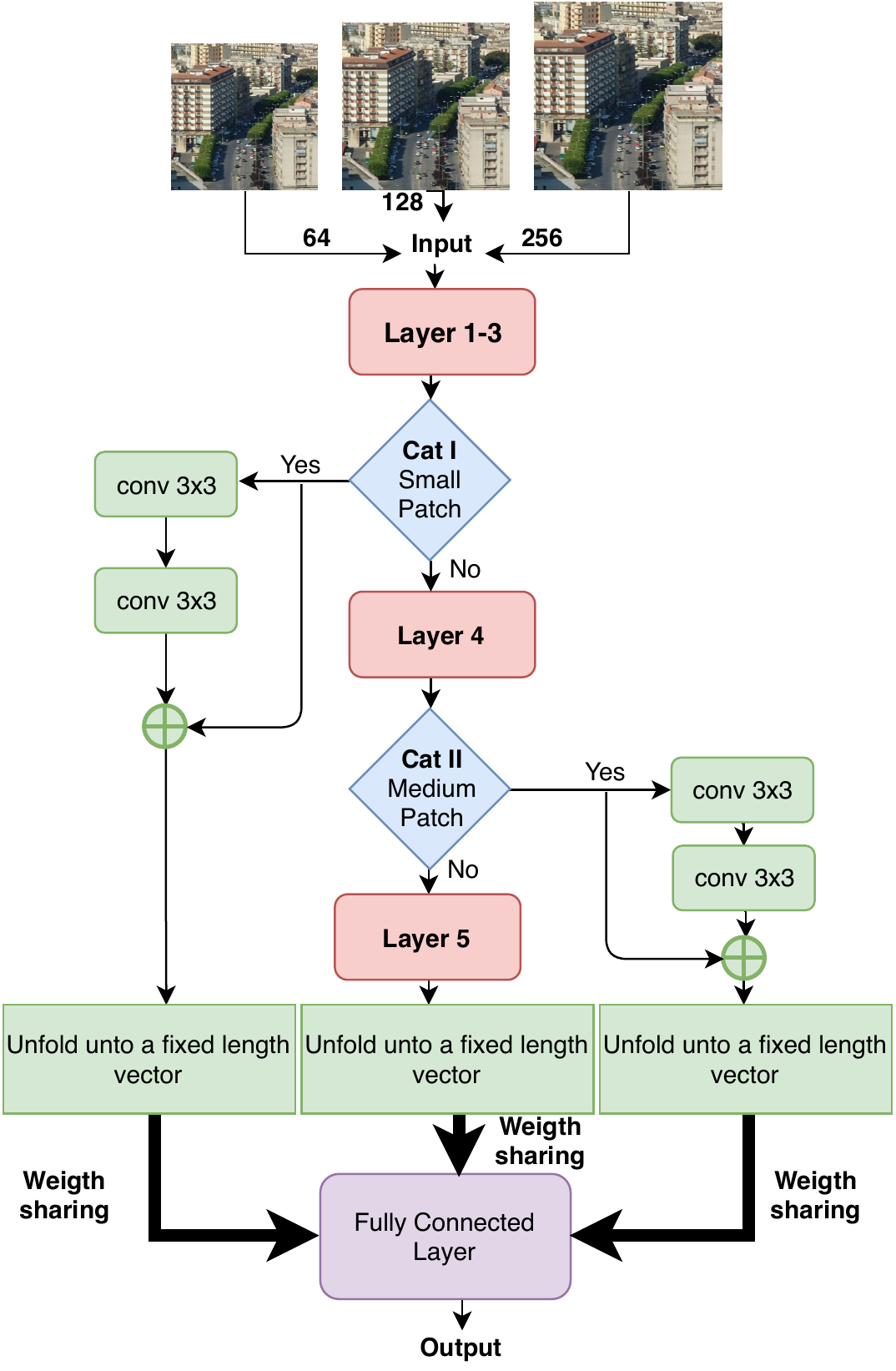}
    \caption{A schematic representation of the proposed \textit{Branched Network (BN)} for handling patches of different sizes with prior information.}
    \label{fig:branchedRESNET}
\end{figure}
Deeper thought on \textit{iterative pooling} suggests that patches of larger size get to see the pooling kernel more number of times. If this recursive relation is unfolded it would result in a branched architecture concerning patches of multiple sizes. There is however a  small issue with the way IPN does this branching which complicates the classification problem. Let's say an image of size $512\times512$ is upsampled by 1.2 and an image of size $1024\times1024$ is downsampled by 0.6 factor. The resulting image size is the same for both cases and so IPN processes them identically putting the burden on fully connected layers for correct classification. The discriminatory power of the network is however enhanced if the two images were processed separately and this is the core idea for \textit{Branched Network (BN)}. BN is given some prior information on image size before resampling and this is used to do the branching as depicted in Fig. \ref{fig:branchedRESNET}. In our experiments this gives better performance than IPN.

Another naive solution is to have a separate CNN for each category. However, inspired by the success of sharing weights of the region proposal network and classification network of \cite{fasterRCNN}, we do not opt for this highly inefficient brute force solution. The idea is that there will be a lot of redundancy in such a system. Ordinarily, the initial layers of a CNN do some basic filtering such as edge extraction which is largely task-independent. Task-specific work is chiefly done by the later layers. Building on this intuition we successively share most
of the layers and keep only one or two size-specific layers. Since size-specific layers are few, they get fine-
tuned in few iterations. The other layers are also benefitted
as they are backpropagated for any resolution and so updated
more frequently helping in faster convergence.

The five layers in Fig. \ref{fig:branchedRESNET} are the five convolutional layers of ResNet-18 \cite{ResidualBlock}. The proposed network, called \textit{Branched Network (BN)} is designed in such a way that the initial layers, which are task-independent, are shared by all patches. As we go deeper, the smaller patches gain sufficient receptive field and branches out. The larger patches continue together deeper till they also achieve an appreciable receptive field. Ultimately, this is followed by a common fully connected layer. The last block just before the fully connected layer consist of  independent max-pooling operation to reduce the spatial resolution to $1\times1$ and one-cross-one convolution to have same number of channels. They are then vectorized and fed to the fully connected layer for classification.

\section{Experiments}
\subsection{Dataset Preparation}
We use the raw image dataset, RAISE \cite{raise}, for experiments on resampling detection. We choose five image sizes: $512\times512$, $1024\times1024$, $3008\times2000$, $4288\times2848$ and $4928\times3264$. We use the PyTorch framework \cite{pytorch} for our implementation and we will make the source code public for future development.

We now describe the procedure for data preparation for image dimension $1024\times1024$. The same procedure was repeated for all other image sizes with exact split and random seed values. An uncompressed image was taken from the RAISE dataset and the central $1024\times1024$ region was cropped. The image was then JPEG compressed in MATLAB by randomly choosing a quality factor from 50 to 100 with a step of 10. It was then resampled by five resampling factors, namely, $0.6, 0.8, 1, 1.2 \text{ and } 1.4$ using the \textit{imresize} command of MATLAB. Each of them was again JPEG compressed at quality factor of $90$. This was repeated for $5000$ images to constitute the training set and a disjoint set of $1000$ images for the testing set. Hence the total training set for image dimension $1024\times1024$ was $25000$ with each resampling factor having $5000$ images. Similarly, the testing set had $5000$ images with each resampling factor having $1000$ images.

For very large image sizes such as $4928 \times 3264$, sometimes the central region could not be cropped and that particular instance was discarded. In this way, with all image sizes, the training set had $75,000$ images with all the five resampling factors having images in equal proportion. Similarly, the testing set of $15,000$ images was equally divided amongst all five resampling factors. Since this is a case of balanced class distribution, detection accuracy for each resampling factor is taken as the total number of correctly classified images. 
%Attempts to hype the accuracy of one class by increasing the false positives would result in an equal dip in accuracy for other classes and therefore not artificially boosting the average accuracy.

\begin{table}[th!]
\centering
\caption{Results on MISLnet \cite{stamm2018TIFS} for JPEG+Res+JPEG. '-' denotes absence. When trained and tested for a fixed image resolution of $1024\times1024$, MISLnet's accuracy surges to $99$\%. However, when trained and tested simultaneously for varying image resolutions, MISLnet is barely able to go past $90$\% accuracy. 
%This is expected because \textit{MISLnet} was not intended to handle different image resolutions. 
}
\begin{adjustbox}{max width=\linewidth,max totalheight=\textheight,keepaspectratio}
\begin{tabular}{c|cccccc}\hline \textbf{Patch size /} & \multicolumn{6}{|c}{\textbf{Resampling Factors}}\\
\cline{2-7} \textbf{Img Resolution} & \textbf{0.6} & \textbf{0.8} & \textbf{1} & \textbf{1.2} & \textbf{1.4} & \textbf{Avg Acc \%} \\ \hline
\textbf{256 / 512} & - & - & - & - & - & - \\
\textbf{256 / 1024} & 99.3 & 98.8 & 99.1 & 99.4 & 99.0 & \textbf{99.0}  \\
\textbf{256 / Larger} & - & - & - & - & - & - \\ \hline
\textbf{256 / 512} & 72.3 & 79.3 & 98.3 & 93.0 & 93.8 & 87.25 \\
\textbf{256 / 1024} & 92.9 & 80.4 & 98.3 & 99.3 & 98.0 & \textbf{93.60} \\
\textbf{256 / Larger} & 89.0 & 73.5 & 97.5 & 96.6 & 93.4 & 89.90  \\ \hline
\end{tabular}
\end{adjustbox}
 \label{Mislnet_compre}
\end{table}

\begin{table}[t!]

\centering
\caption{Results for resampling factor detection on double JPEG compressed images. All the listed solutions gave close to $99\%$ accuracy when trained and tested on images of fixed dimension but dropped by $10\%$ when retrained and tested for images of varying dimensions. The proposed solutions, IPN and BN help to regain the lost accuracy. Accuracy in \%.}
\begingroup
\setlength{\tabcolsep}{4pt}
\begin{tabular}{c|cccccc}
\hline \textbf{Method}  & \multicolumn{5}{c}{\textbf{Resampling Factors}} &      \\
 \cline{2-7}       & \textbf{0.6}   & \textbf{0.8}    & \textbf{1}     & \textbf{1.2}   & \textbf{1.4}   & \textbf{Avg}  \\
\hline MISLnet \cite{stamm2018TIFS} & 84.7  & 77.7   & 98.0  & 96.3  & 94.9  & 90.3 \\
Quan \etal~\cite{tifs2018prcg}    & 80.4  & 78.9   & 97.3  & 97.1  & 95.8  & 89.9 \\
Chen \etal~\cite{2019_multi}    & 73.4  & 79.13  & 97.3  & 94    & 94.9  & 87.7 \\
       \hline
MPN     & 78.3  & 48.0   & 93.7  & 52.3  & 91.1  & 72.7 \\
IPN (ours)   & 98.6  & 97.1   & 98.4  & 94.3  & 94.8  & \textbf{96.6} \\
BN (ours)     & 98.4  & 99.0   & 98.0  & 98.8  & 99.5  & \textbf{98.7} \\ \hline
\end{tabular}
\endgroup
\label{table:mainn}
\end{table}
\begin{table}[t!]
\centering
\caption{Patch sizes for the proposed IPN and BN.}
\begin{tabular}{c|cc}
\hline
\textbf{Image size} & \multicolumn{2}{c}{\textbf{Patch size $p\times p$}} \\ 
$d\times d$         & IPN                       & BN                       \\ \hline
$d<1024$            & 128                       & 64                       \\ 
$1024<d<2000$       & 256                       & 128                      \\ 
$d>2000$            & 512                       & 256                      \\ \hline
\end{tabular}
\label{patchsize}
\end{table}
\begin{table}[t!]
\centering
\caption{ Patch level accuracy in \% for Natural Images (NI) vs. Computer Generated (CG) images of heterogeneous origin on the Columbia dataset \cite{prcg_dataset}.}
\begin{tabular}{c|ccc}
\hline \textbf{Method} & \textbf{Google NI} & \textbf{Personal NI} & \textbf{CG}  \\ \hline
{Quan \etal~ \cite{tifs2018prcg}} & 75.8 & 88.7 & 68.6  \\
{{MISLnet} \cite{stamm2018TIFS}}& 70.2 & 70.3 & 49.0 \\ 
{IPN (Ours)} & \textbf{90.0} & \textbf{94.0} & \textbf{78.0}  \\ \hline
\end{tabular}

\label{prcg_table}
\end{table}
\subsection{Results on Resampling Detection}
The results for resampling detection in double compressed JPEG images are shown in Table \ref{table:mainn}. When the previous works listed in the table were re-trained on images of fixed base dimension of $1024\times1024$ with patch chosen from image center, the following average accuracies were obtained \textemdash~ MISLnet \cite{stamm2018TIFS}: $98.7\%$, Quan \etal~ \cite{tifs2018prcg}: $98.4\%$ and Chen \etal~ \cite{2019_multi}: $98.2\%$. However, interestingly when they were re-trained on all image sizes about $10\%$ drop in average accuracy can be observed in Table \ref{table:mainn}. Since these works cannot alter patch size, the default patch size chosen for the architecture was used. It was in the range of $200-300$. 
The table next reports the average accuracy obtained by using the adaptive max-pooling operation instead of iterative pooling in IPN. This is referred as the Max-Pooling Network (MPN). As suggested in Sec. \ref{sec: iterative pooling}, it is unable to restore the accuracy. IPN on the other side has very promising results, restoring much of the lost accuracy. Also BN with the help of prior information about the original image size again gives close to $99\%$ average accuracy and with much smaller patch sizes. The choice of patch size for IPN and BN is listed in Table \ref{patchsize}.

\begin{figure}
    \centering
    \includegraphics[width=1\linewidth]{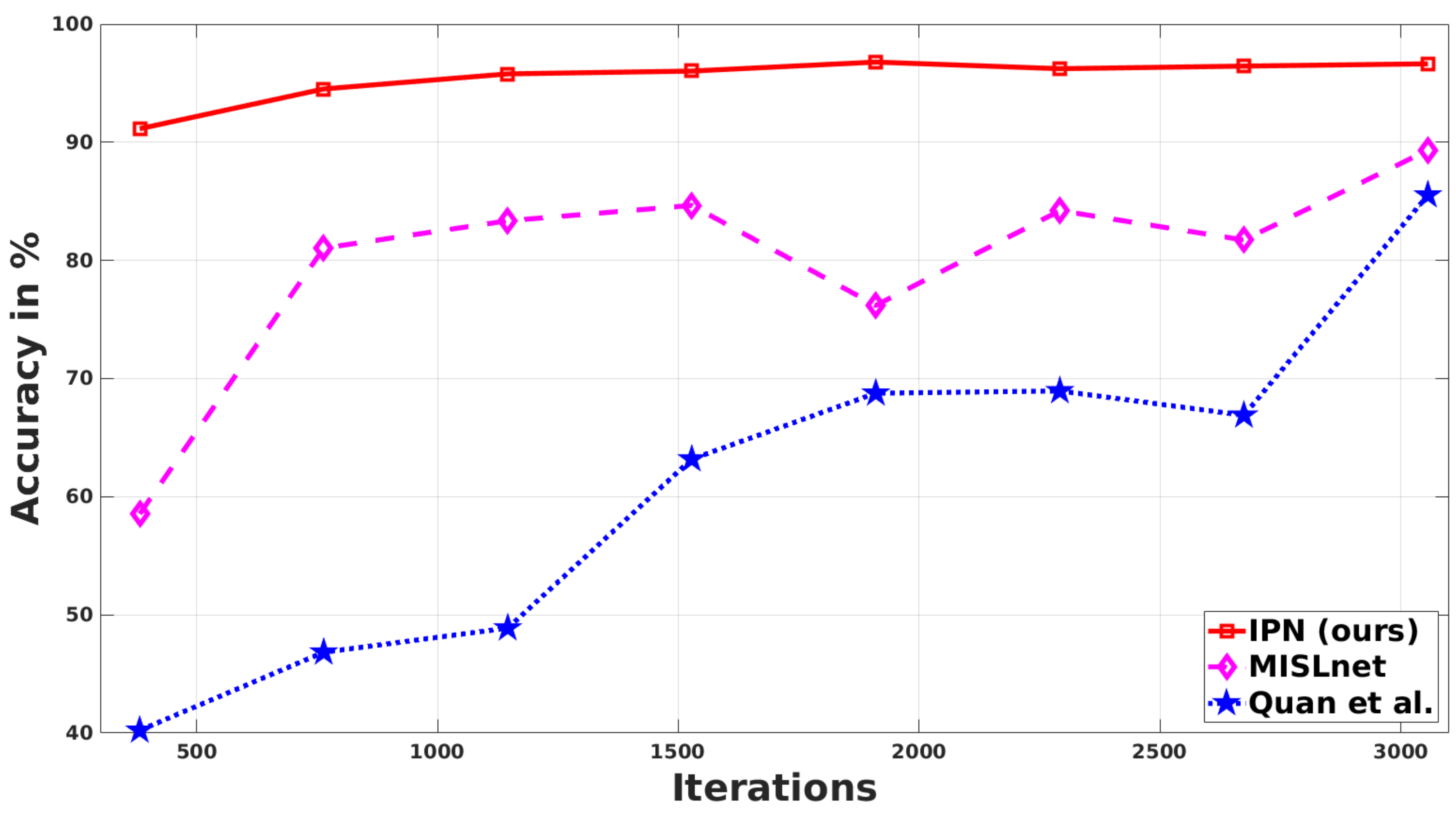}
    \caption{The proposed IPN  also converges faster (in about $2000$ iterations) for resampling detection task with final accuracy reported in Table \ref{table:mainn}.  The other two methods, MISLnet \cite{stamm2018TIFS} and Quan \etal~\cite{tifs2018prcg} took up to 6000 iterations to stabilize. The curve for Chen \etal~\cite{2019_multi} not shown because of extremely slow convergence taking 15000+ iterations to converge.}
    \label{fig:convergence}
\end{figure}

\subsection{Results on Photorealism Detection}
The flexibility in the choice of patch size is the main strength of the proposed solutions. In this section, we describe some of the added benefits that are obtained by building on top of pre-trained models.
%unlike the recent works such as \cite{tifs2018prcg, stamm2018TIFS}. 
Firstly, the convergence for the resampling detection task is much faster, as seen in Fig. \ref{fig:convergence}. Secondly, the CNN model proposed in \cite{2017stamm,stamm2018TIFS} are better suited for resampling detection task and a few other tasks, while the one proposed in \cite{tifs2018prcg} is better suited for photorealism detection. However, by building on top of already existing state-of-the-art architectures helps easy generalization to different problems. For example, by using ResNet-18 \cite{ResidualBlock} as base architecture, IPN achieves better performance for both resampling and photorealism detection tasks as corroborated by Table \ref{table:mainn} and Table \ref{prcg_table}. For Table \ref{prcg_table}, the exact procedure followed by \cite{tifs2018prcg} on the Columbia dataset \cite{prcg_dataset} was used to re-train the listed CNN models, except for one change. The Natural Images (NI) and Computer Generated Images (CG) were not resized to a fixed dimension before patch selection. This is to prevent the CNNs from classifying based on the resampling factor. Similarly we test the proposed method for JPEG+Res+Rot+JPEG forgery. For this set up $1024 \times 1024$ images were rotated clockwise and anti-clockwise in the range -20 degrees to +20 degrees. But for detection only clockwise or anti-clockwise rotation has to be detected. This restriction is made because this is a much harder problem than the one mentioned in Table 3. Rest of the settings remain the same.

\begin{table}[]
\centering
\caption{Performance of IPN on JPEG+Res+Rot+JPEG. Rot Clock: Clockwise rotation; Rot Anti: Anti-clockwise rotation; Res Acc: Resampling detection accuracy; Rot Acc: Rotation detection Accuracy; Res 0.6: Resampling by factor of 0.6.}
\begin{adjustbox}{max width=\linewidth,max totalheight=\textheight,keepaspectratio}
\begin{tabular}{cccccc}
\hline \textbf{} & \textbf{} & \textbf{} & \textbf{IPN} & \textbf{\textit{MISLnet}} & \textbf{Quan} \\ &&&(\textbf{ours})&\textbf{\cite{stamm2018TIFS}}&\textbf{et al.} \textbf{\cite{tifs2018prcg}}\\ \hline
\multirow{10}{*}{\textbf{Rot Anti}} & \multirow{2}{*}{\textbf{Res 0.6}} & Res Acc & 96.6 & 83.3 & 77.9 \\
 &  & Rot Acc & 95.2 & 85.2 & 38.9 \\ \cline{2-6}
 & \multirow{2}{*}{\textbf{Res 0.8}} & Res Acc & 93.9 & 75.5 & 51.3 \\
 &  & Rot Acc & 95.3 & 82.7 & 56.8 \\ \cline{2-6}
 & \multirow{2}{*}{\textbf{Res 1}} & Res Acc & 95 & 84.8 & 62.2 \\
 &  & Rot Acc & 99.1 & 86.9 & 66.5\\ \cline{2-6}
 & \multirow{2}{*}{\textbf{Res 1.2}} & Res Acc & 96.6 & 89.2 & 52.6 \\
 &  & Rot Acc & 98.1 & 87.5 & 76.5\\ \cline{2-6}
 & \multirow{2}{*}{\textbf{Res 1.4}} & Res Acc & \multicolumn{1}{c}{98.9} & \multicolumn{1}{c}{96.5} & 85.4\\
 & \multicolumn{1}{l}{} & Rot Acc & 99.3 & 91.7 & 75.0\\ \hline
 \multirow{10}{*}{\textbf{Rot Clock}} & \multirow{2}{*}{\textbf{Res 0.6}} & Res Acc & 96.4 & 82.1 & 78.1\\ 
 &  & Rot Acc & 94.4 & 85.5 & 68.0\\ \cline{2-6}
 & \multirow{2}{*}{\textbf{Res 0.8}} & Res Acc & 95 & 75.4 & 50.5\\
 &  & Rot Acc & 94.8 & 86.9 & 51.0\\ \cline{2-6}
 & \multirow{2}{*}{\textbf{Res 1}} & Res Acc & 94.9 & 83.8 & 61.8\\
 &  & Rot Acc & 98.7 & 91.6 & 38.6\\ \cline{2-6}
 & \multirow{2}{*}{\textbf{Res 1.2}} & Res Acc & 96.4 & 90.3 & 52.7\\
 &  & Rot Acc & 98.3 & 93.5 & 26.9\\ \cline{2-6}
 & \multicolumn{1}{l}{\multirow{2}{*}{\textbf{Res 1.4}}} & \multicolumn{1}{l}{Res Acc} & \multicolumn{1}{c}{99.1} & \multicolumn{1}{c}{97.5} & 87.0\\
 & \multicolumn{1}{l}{} & Rot Acc & 99.4 & 94.6 & 27.4\\ \hline
 \multicolumn{3}{c}{\textbf{Resampling Avg Acc \%}} & \textbf{96.2} & 85.8 & 66.0\\
\multicolumn{3}{c}{\textbf{Rotation Avg Acc \%}} & \textbf{97.2} & 88.6 & 52.5\\ \hline
\end{tabular}
\end{adjustbox}
\label{rot_table}
\end{table}

%This way the CNNs may learn to differentiate the resampling factors rather than focussing on the real task because NI and CG have quite different image resolutions.

\section{Conclusion}
Many image forensic problems require to alter the patch size depending on the image size. However, the fully connected layers of CNNs do not allow this. To tackle this issue, we proposed two solutions: IPN and BN. IPN converts variable size tensors to a fixed dimension by iteratively convolving them and BN had dedicated pathways for different image sizes. This helped to restore 7-9\% accuracy for resampling detection which was lost by exposing existing solutions to multiple image sizes. We also demonstrated that by building on top of existing state-of-the-art deep learning models such as ResNet18, we obtained stable iterations and faster convergence. This approach also generalizes well for other tasks such as photorealism detection.
%Different image acquisition devices capture images of different sizes and therefore, in today's time, the inability of an algorithm to work on images of varying dimensions limits its scope. Images of multiple sizes many times requires to alter the patch size, which unfortunately is difficult with CNNs having fully connected (fc) layers. Even for the task of resampling detection in double compressed images, a simple ubiquitous operation for most image manipulations, our experiments showed that exposing the recent image forensic methods to images of multiple sizes caused a significant dip in accuracy. We, therefore, proposed the \textit{iterative pooling} strategy, that can be integrated with CNNs having fc layers to handle varying patch sizes. This strategy does not require any prior information and gave significant improvement. If, however, some prior is provided, the proposed \textit{Branched Network} gave even better results. The idea of building on existing pre-trained models was also emphasized to get faster convergence and easy parameter tuning by showing results on resampling and photorealism detection, which have been to date solved using models of different choices.
\bibliographystyle{IEEEbib}
\bibliography{strings,refs}

\end{document}